\documentclass{article}
\usepackage{ijcai23}

\usepackage{times}
\usepackage{soul}
\usepackage{url}
\usepackage[hidelinks]{hyperref}
\usepackage[utf8]{inputenc}
\usepackage[small]{caption}
\usepackage{graphicx}
\usepackage{amsmath}
\usepackage{amsthm}
\usepackage{booktabs}
\usepackage{algorithm}
\usepackage{algorithmic}
\usepackage[switch]{lineno}

\usepackage{soul}
\usepackage{url}
\usepackage[utf8]{inputenc}
\usepackage{graphicx}
\usepackage{amsmath}
\usepackage{xcolor}
\usepackage{booktabs}
\usepackage{multirow}
\usepackage{amsmath,xparse,mleftright}
\usepackage{amssymb}
\usepackage{caption}

\usepackage{amsthm}

\usepackage{caption}
\usepackage{mathrsfs}

\title{RZCR: Zero-shot Character Recognition via Radical-based Reasoning}

\author{
Xiaolei Diao\textsuperscript{1,2}\and
Daqian Shi\textsuperscript{2} \and 
Hao Tang\textsuperscript{3} \and
Qiang Shen\textsuperscript{1} \and
Yanzeng Li\textsuperscript{4} \and
Lei Wu\textsuperscript{4} \and
Hao Xu\textsuperscript{1\footnote{Corresponding Author}}
\affiliations
\textsuperscript{1}College of Computer Science and Technology, Jilin University\\
\textsuperscript{2}DISI, University of Trento\\
\textsuperscript{3}CVL, ETH Zurich\\
\textsuperscript{4}College of Software Engineering, Jilin University
\emails
\{xiaolei.diao, daqian.shi\}@unitn.it,
hao.tang@vision.ee.ethz.ch,\\
\{shenqiang19, yzli20, wulei20\}@mails.jlu.edu.cn,
xuhao@jlu.edu.cn
}
\begin{document}

\maketitle

\begin{abstract}
The long-tail effect is a common issue that limits the performance of deep learning models on real-world datasets. Character image datasets are also affected by such unbalanced data distribution due to differences in character usage frequency. Thus, current character recognition methods are limited when applied in the real world, especially for the categories in the tail that lack training samples, e.g., uncommon characters. In this paper, we propose a zero-shot character recognition framework via radical-based reasoning, called RZCR, to improve the recognition performance of few-sample character categories in the tail. Specifically, we exploit radicals, the graphical units of characters, by decomposing and reconstructing characters according to orthography. RZCR consists of a visual semantic fusion-based radical information extractor (RIE) and a knowledge graph character reasoner (KGR). RIE aims to recognize candidate radicals and their possible structural relations from character images in parallel. The results are then fed into KGR to recognize the target character by reasoning with a knowledge graph. We validate our method on multiple datasets, and RZCR shows promising experimental results, especially on few-sample character datasets.
\end{abstract}

\section{Introduction}
\label{sec:Introduction}
\begin{figure}[!t]
	\centering
    \includegraphics[width=1\linewidth]{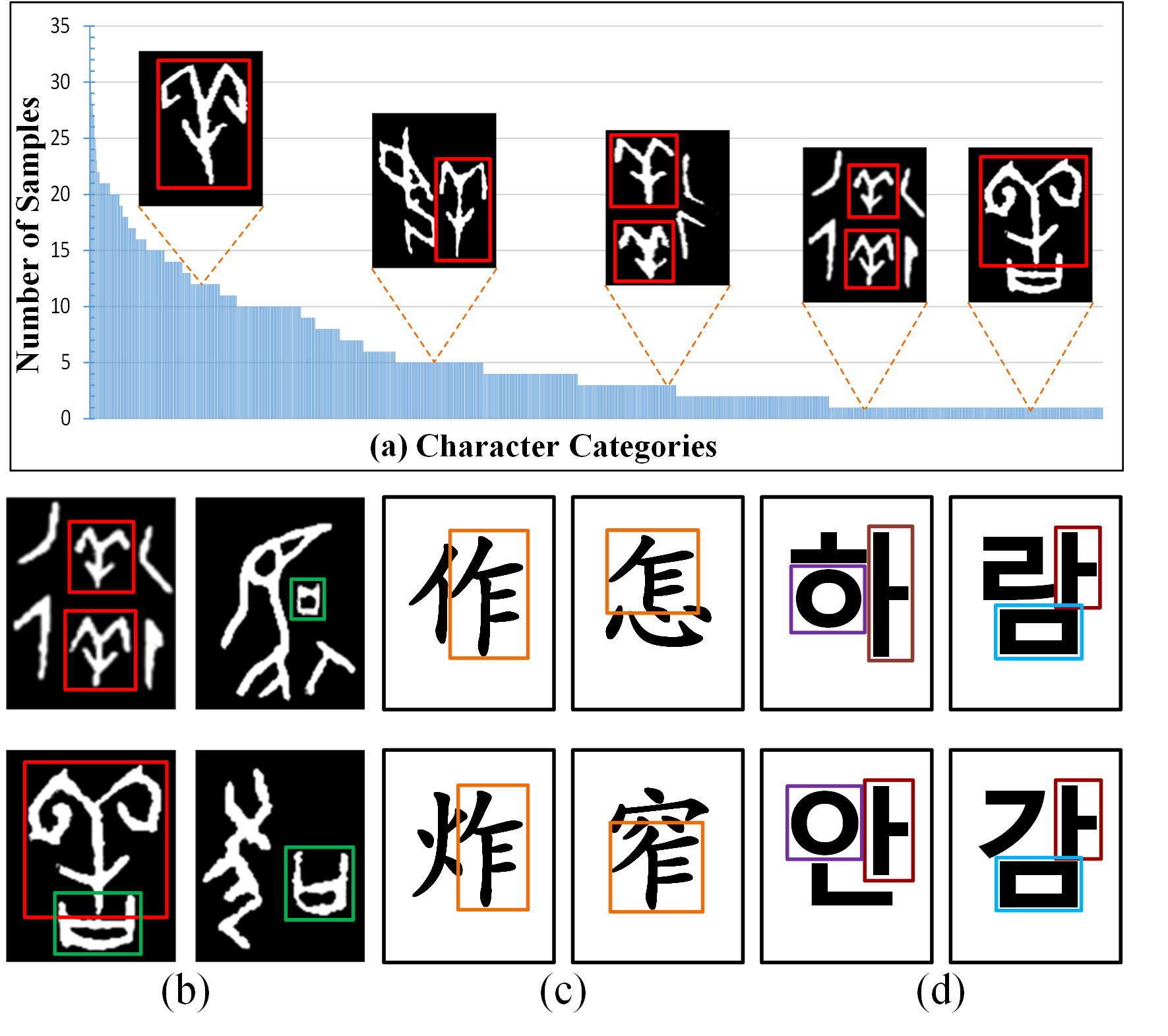}
	\caption{Analysis of characters, where radicals are distinguished by colored boxes. (a) Statistics of a character dataset; (b)-(d) Examples of oracle bone, Chinese, and Korean characters.
	\label{fig:1}}
\end{figure}

Developments in optical character recognition (OCR) technology offer new solutions for learning, managing, and utilizing character resources. Current OCR methods are mainly based on deep learning models, which place high demands on the quantity and quality of data \cite{zhang2017online}. Due to differences in character usage frequency, the long-tail effect exists as a common issue in character datasets. Such datasets often contain categories with few samples, especially for unreproducible or inaccessible cases, e.g., historical character \cite{huang2019obc306} and calligraphic character \cite{lyu2017auto} datasets. Fig.~\ref{fig:1}(a) shows the distribution of character samples for each category in an oracle bone dataset \cite{wu2012}. The distribution demonstrates a typical long-tail effect, where over half of the character categories have five or fewer samples, and there are even categories with a single sample. As a result, such datasets challenge the validity of current OCR methods.

Inspired by studies on general image classification, current research on character recognition mainly focused on methods based on deep convolutional neural networks (CNN). Such OCR methods achieve limited performance on unbalanced datasets, especially when recognizing character categories with few samples \cite{anderson2006long,cao2020zero}. 
A common solution to alleviate the limitations caused by the few-sample problem is employing data augmentation methods to balance data categories, including re-sampling and re-weighting. However, these methods are rarely applied to data categories with few samples or one sample. Although the compromise solution tends to discard categories with few samples \cite{geron2019hands}, we consider such categories valuable and irreplaceable for real-world tasks. For example, many categories in Fig.~\ref{fig:1}(a) contain only one sample in the unearthed oracle bone dataset, which is unique evidence for understanding this character by archaeologists. Thus, we aim to achieve character recognition on few-sample datasets while keeping data integrity.

The orthography-based strategy of learning East Asian characters in terms of radicals and character structure is the most frequent and effective way for human beings \cite{shen2005investigation}. We observe the orthographic peculiarities \cite{myers2016knowing} of East Asian characters:
(1) Characters are composed of a set of radicals organized in a specific structure, where radicals are semantic units of the character.
(2) Radicals are commonly shared by different characters, thus, the number of radicals is less than that of characters significantly. 
Fig.~\ref{fig:1}(b), (c) and (d) show our observations on oracle bone, Chinese, and Korean characters, where we highlight radicals by colored boxes. Statistically, 6,763 Chinese characters can be represented by 485 radicals and 10 character structures \cite{liu2011casia}. In Fig.~\ref{fig:1}(a), we demonstrate some characters that contain the same radical ``yang'' (in red boxes), where such characters tend to distribute randomly in the statistics. As a result, decomposing and reconstructing characters by exploiting radicals help to intensionally analyze characters, which offers the possibility to learn from enough radical samples and recognize character categories with few samples. 

According to the above-mentioned discussions, we propose a novel method for zero-shot character recognition via radical-based reasoning, namely RZCR. Specifically, the proposed RZCR consists of a radical information extractor (RIE) and a knowledge graph-based character reasoner (KGR). 
The former aims to extract visual semantic information in characters to obtain candidate radicals and their structural relations in parallel through the radical attention blocks (RABs) and structural relation blocks (SRBs), respectively. 
A fused attention layer is proposed to fuse visual semantic information extracted by RABs and SRBs, and a dual spatial attention layer in RABs is introduced to handle the radical overlapping issues. 
The results of RIE are then fed into KGR, where a weight-fusion reasoning algorithm is proposed to reconstruct characters by modeling the radical information as soft labels. We achieve zero-shot recognition by reasoning with the pre-designed character knowledge graph (CKG) that stores information about characters, radicals, and their relations.

The contributions of our paper are summarized as follows:

\begin{itemize}
    \item We propose a novel zero-shot method (i.e., RZCR) for character recognition, which can effectively handle categories with insufficient training samples by character decomposing and reconstructing. 
     \item RZCR introduces a new strategy to organize candidates for reasoning-based character recognition, where candidate radicals and structural relations are extracted in parallel by RIE, and then characters are recognized by the proposed weight-fusion reasoning algorithm in KGR.
    \item Our method is validated on multiple datasets, including a newly constructed dataset, namely OracleRC. Compared to state-of-the-art OCR methods, our RZCR achieves promising results, especially on few-sample cases.
\end{itemize}

\section{Related Work}
\label{sec:Related Work}
\noindent\textbf{Zero-shot Learning.}
Zero-shot learning aims to classify unseen categories by learning from existing categories and knowledge supplemented by auxiliary resources \cite{wang2019survey,sun2021research}. In recent years, zero-shot learning has been considered from a variety of perspectives. For instance, embedding-based methods \cite{akata2015label,xie2019attentive} propose to perform image embedding and label semantic embedding in the same space and extend to unseen categories by a compatibility function. Furthermore, attribute-based methods \cite{lampert2009learning,lampert2013attribute} manually design attributes in different ways and represent categories with multidimensional attribute vectors, which are introduced to train the visual classifier by vector mapping. Moreover, reasoning-based methods exploit knowledge graphs \cite{rohrbach2011evaluating,wang2018zero} to guide zero-shot learning since knowledge graphs connect trained categories and unseen categories by pre-defined relations. The success of the above methods brings us inspiration about the usage of auxiliary information from knowledge graphs and the organization of attributes in zero-shot character recognition. 

\noindent\textbf{Character Recognition.}
Early studies on character recognition mainly contain feature extraction-based \cite{su2003novel} and statistical-based methods \cite{shanthi2010novel}, which perform poorly on datasets with a large number of categories. Then, researchers exploit CNN, e.g., \cite{yuan2012offline} applies an improved LeNet-5 model to recognize English characters, and \cite{cirecsan2015multi} first introduces CNN into Chinese character recognition. \cite{zhang2017online} combines a normalization-cooperated direction-decomposed feature map with deep CNN for handwriting character recognition. To recognize curved or distorted characters in the real world, \cite{zhan2019esir,yang2019symmetry} introduce trainable models based on symmetry-constrained rectification and line-fitting transformation, respectively. Moreover, deep character recognition models with dedicated improvements achieve promising results in different languages \cite{balaha2021new,mushtaq2021urdudeepnet}. 
% \noindent \textbf{Radical-based Methods.}
The success of the above deep learning-based methods relies on a large number of training samples for each character category. Thus, data augmentation strategies are introduced to solve insufficient data issues. For instance, \cite{qu2018data} combines global transformation and local distortion to effectively enlarge the training dataset, and \cite{Luo_2020_CVPR} designs a set of custom fiducial points to assist in flexibly enhancing character images, both of which alleviate the insufficient training sample issue. Some researchers also introduce domain knowledge related to character attributes to deal with few-sample problems by decomposing characters, i.e., presenting characters as preset sequences. \cite{cao2020zero,zhang2020radical} introduce CNN-based encoder-decoder frameworks to generate radical sequences for character recognition. Inspired by these methods, \cite{chen2021zero} decomposes characters into strokes, the smallest units of characters, then perform recognition by looking up generated stroke sequences in a dictionary. However, the above methods require the generated sequences to match the dictionary exactly, both for each element and the order, i.e., hard-matching strategy. As a result, such sequence-based methods perform poorly on the benchmarks, which limits their applications in practice.

\section{Intuitive Discussion}
\label{sec:Intuitive Discussion}

In this section, we provide some intuitive observations to explore the character recognition task and discuss potential performance improvement strategies, driving the motivation of this paper. Then, we model the zero-shot character recognition to clarify the task in this paper.

\begin{figure}[!t]
	\centering
	\includegraphics[width=1\linewidth]{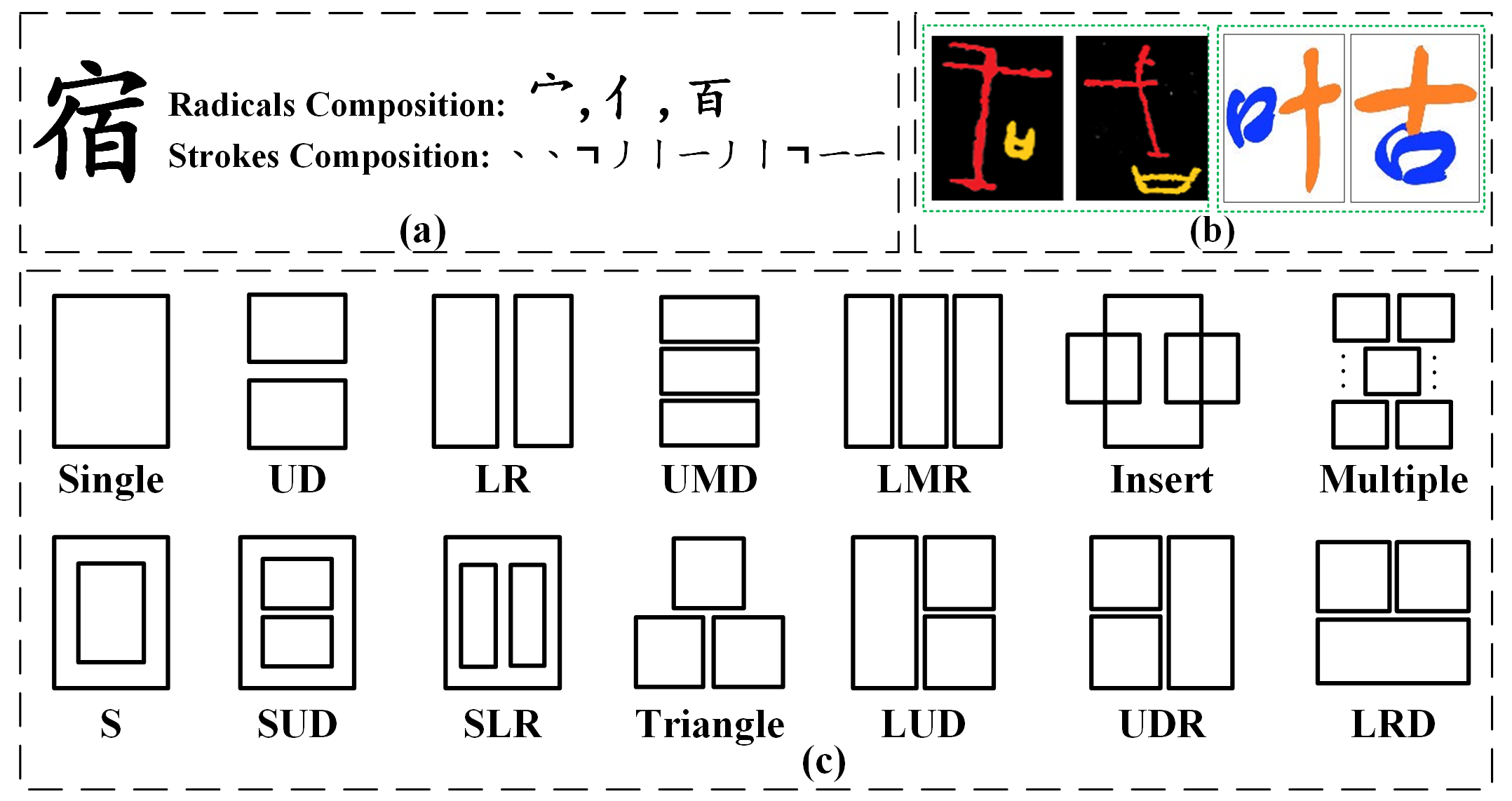}
	\caption{(a) Character composition in radical and stroke level, respectively; (b) Examples of characters composed by same radicals (highlighted in same colors) but different structural relations; (c) 14 predefined structural relations. 	
	\label{fig:2}}
\end{figure}

\noindent\textbf{Decomposition of Characters.}
According to orthography, characters can be decomposed into a set of radicals organized in specific structures, where radicals are independent characters or evolve from simple-semantic characters \cite{yeung2016orthographic}. Complex-semantic characters usually contain more than one radical since radicals are the smallest units that represent complete semantic information \cite{ho2003radical}. A Chinese character is shown in Fig.~\ref{fig:2}(a) as an example, where the three radicals in the first row represent ``house'', ``person'', and ``mat (evolved)'' and constitute the character that means ``get an accommodation''. However, the decomposition in the second row shows an opposite example, i.e., the character is decomposed excessively into strokes, where the major semantic information is lost. 
Meanwhile, we define the structure of character as the relative localization relation between the constituent radicals, namely structural relations, which also contain semantic information. Structural relations are also crucial for characters since the same radical composition points to different candidate characters when the structure changes, as shown in Fig.~\ref{fig:2}(b). 
% 
% Meanwhile, we define the structure of character as the relative localization relation between the constituent radicals, namely "structural relations", which also contain semantic information. Structural relations are also crucial for characters since the same radical composition will point to different candidate characters when the structure changes. Two sets of examples are given in Fig.~\ref{fig:2}(b), where different characters in the red boxes contain the same radical composition but different structural relations. 
% 
Moreover, as discussed earlier, the same radical can be shared by various characters, e.g., 2,374 Korean characters contain 68 radicals \cite{zatsepin2019fast}, and over 6,000 common Chinese characters can be composed of less than 500 radicals \cite{liu2011casia}, which means a large number of characters can be presented by a small number of radicals.

\noindent\textbf{Motivation and Challenges.}
% As a result, two ideas are derived from the above observations for achieving zero-shot character recognition. Firstly, a character can be properly decomposed into a set of radicals and their structural relations, i.e., radical information, which can be extracted from the character image. Since the semantics of characters can be represented by radical information, we aim to extract such semantic information through deep neural networks. Thus, we propose RIE to extract all the radical information in parallel rather than extracting separately since radicals and the structure support each other. Meanwhile, we apply the attention mechanism to deal with the radical overlapping problem for better recognition performance.
Based on the above observations, two ideas are derived for achieving zero-shot character recognition. Firstly, a character can be properly decomposed into a set of radicals and their structural relations, i.e., radical information, which can be extracted from the character image. Since the semantics of characters can be represented by radical information, we aim to extract such semantic information through deep neural networks. Thus, we propose RIE to extract all the radical information in parallel rather than extracting separately since radicals and the structure support each other. Meanwhile, we apply the attention mechanism to deal with the radical overlapping problem for better recognition performance.

% Note that, extracting radical categories from character images require localizing radicals, which will help determine the structural relationship between radicals. 

Secondly, current sequence-based methods organize character elements in the form of long sequences that are matched by hard-matching strategies. In Table \ref{tab:ac-1}, we show accuracy estimation for sequence-based methods at both radical and stroke levels, where $p$ represents the probability of correct character recognition; $r_i, s_i$ are correct recognition probabilities of elements in radical and stroke sequences, respectively; $m, n$ are the lengths of sequences. We find $p$ only achieves 0.48/0.28 maximally under the hard-matching strategy, even if we ideally assume a high recognition probability $r_i {=} s_i {=} 0.9$. Therefore, we propose KGR to achieve reasoning-based zero-shot recognition by exploiting knowledge graphs to organize characters, radicals, and structural relations in a flexible way rather than in long sequences. Meanwhile, we introduce soft labels into the matching strategy, i.e., the proposed weight-fusion algorithm, to achieve stable recognition performance. 

\begin{table}[!t]
\centering
\resizebox{1\linewidth}{!}{
\begin{tabular}{@{}l|ccc@{}}
\toprule
\textbf{Method} & \textbf{ASL} & \textbf{Accuracy Calculation} & \textbf{Example ( $r_i,s_i=0.9$)} \\ \midrule
Radical level    & 7.76                    & $p = \prod \limits_{i=0}^m r_i$
                  & ${0.9}^7 = 0.4783$          \\
Stroke level     & 12.88                   & $p = \prod \limits_{i=0}^n s_i$                  & ${0.9}^{12} = 0.2824$   \\ \bottomrule
\end{tabular}
}
\caption{Accuracy estimation for sequence-based methods. ASL refers to the average sequence length.}
\label{tab:ac-1}
\end{table}

\noindent\textbf{Task Formulation.}
The goal of the character recognition is to classify character image $I_{C}{\in}\mathbb{R}^{H{\times} W{\times} C}$ into character category $Z$, where $H{\times} W$ represents the spatial resolution, and $C$ is the number of channels. We model character recognition as a zero-shot classification task, defined as follows. The training data is defined as $I_{seen}{=} \{(x, y)| x {\in} X, y {\in} Y\}$, where $x$ is the input image from the training image set $X$, $y$ refers to the label(s) of $x$ annotated based on radical information, and $Y$ is the label set of the radical categories. Note that there are one or more radicals in an image. Similarly,  the test set can be defined as $I_{unseen} {=} \{(x, z)|x {\in} \hat{X}, z {\in} Z\}$, where $x$ belongs to images from unseen categories $\hat{X}$, i.e., character categories, $Z$ represents the set of labels for unseen categories. In this task, the training is based on radical categories and finally outputs the target character category as $f {:} x {\rightarrow} Z$.

\section{The Proposed RZCR}
\label{sec:Method}
\begin{figure*}[tbp]
	\centering
	\includegraphics[width=1\linewidth]{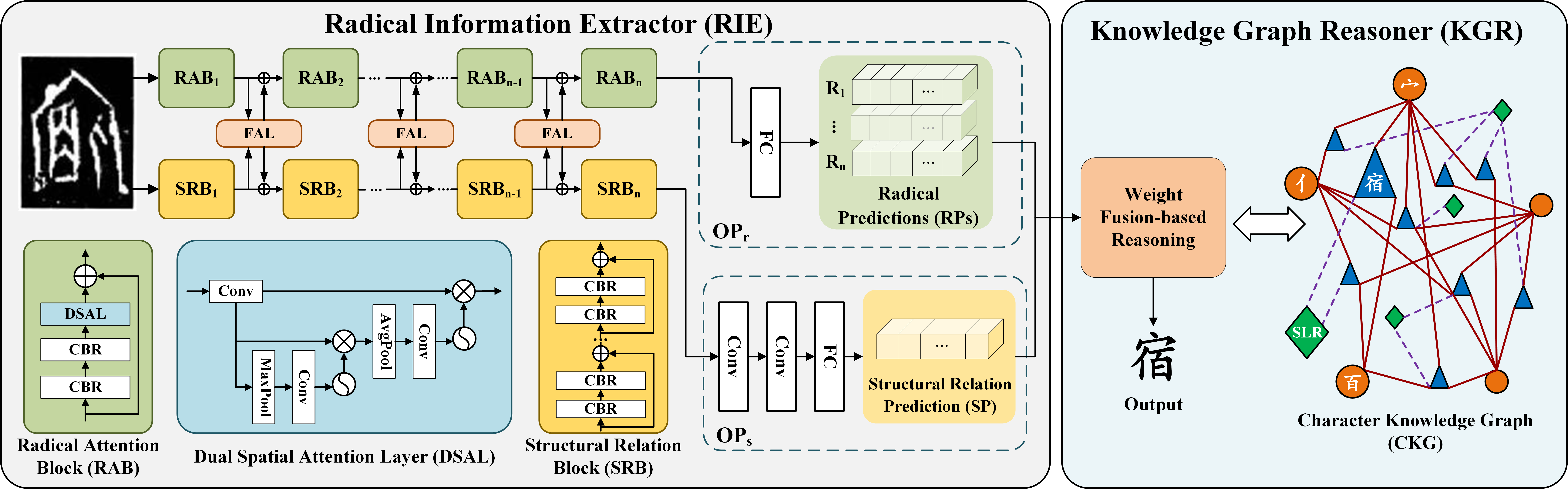}
	\caption{The overall architecture of RZCR. (left) RIE extracts candidate radicals and structural relations from input character images; (right) KGR recognizes the target character by reasoning with the CKG, where blue, orange, and green nodes represent characters, radicals, and structural relations, respectively, and lines in CKG represent different relations between nodes.
	\label{fig:3}}
\end{figure*}

In this section, we detail the proposed RZCR and its key components RIE and KGR, as shown in Fig.~\ref{fig:3}.

\subsection{Radical Information Extractor}

The radicals and the structural relations in characters jointly convey character semantics, which shows both of them are indispensable for character recognition. RIE introduces two sets of blocks, Radical Attention Blocks (RABs) and Structural Relation Blocks (SRBs), aiming to obtain candidate radicals and structural relations from an input character image $I_{C}$, respectively. 
We observe that character images usually contain overlapping and unclear boundaries between radicals \cite{shi2022rcrn}, especially for handwritten characters, which challenges the performance of radical recognition. Thus, the RABs apply the attention mechanism to concentrate on the target radicals. RAB consists of a dual spatial attention layer (DSAL) and two consecutive CBR blocks (including a $3{\times}3$ convolutional layer, a batch normalization operation, and a ReLU operation), and SRB consists of a set of residual CBR blocks, as shown in the Fig.~\ref{fig:3}. For RAB, each DSAL obtains the attention weights by two calculations, in which the foreground information is retained for radical feature selection via MaxPool, and the background information is processed via AvgPool to maintain integrity. The process in DSAL can be defined as:
\begin{equation}
\begin{aligned}
    A^{\prime}_i &= \sigma ({\rm Conv}({\rm MaxPool}(F_i))) \otimes F_i, \\
    A_i  & = \sigma ({\rm Conv}({\rm AvgPool}(A^{\prime}_i))) \otimes F_i,
\end{aligned}
\label{equ:1}
\end{equation}
where $A^{\prime}_i$ and $A_i$ denote the intermediate and final output features of the DSAL in the $i{-}th$ RAB; $\sigma$ refers to the sigmoid function; ${\rm Conv}$ refers to the convolutional layer; $F_i$ is the feature map fed into the corresponding DSAL. 

% A fused attention layer (FAL) is proposed to fuse visual semantic information extracted by RABs and SRBs, aiming to extract robust features of radicals and structural relations by mutually supporting each other.
A fused attention layer (FAL) is proposed to fuse visual semantic information extracted by RABs and SRBs, aiming to improve the performance of radical prediction by injecting structural information and position features into RABs. 
In FAL, we introduce channel attention and spatial attention to provide efficient tokens from selective feature maps and to enhance dependency extraction on spatial axes, respectively, which is inspired by \cite{shi2022charformer}. Given an input feature $F_k$, the output of fused attention layer $FAL(F_k)$ is:
\begin{equation}
      FAL(F_k) = M_s(M_c(F_k)+F_k)+M_c(F_k)+F_k,
\label{equ:11}
\end{equation}
where $M_c, M_s$ are channel and spatial attention, respectively. We denote the $i$-th feature $F_{RA}$ as:
\begin{equation}
      F_{RA_i} = FAL(F_{R_i}+F_{S_i})+F_{R_i},
\label{equ:12}
\end{equation}
The $i$-th structural relation feature $F_{SR}$ is expressed as:
\begin{equation}
      F_{SR_i} = FAL(F_{R_i}+F_{S_i})+F_{S_i},
\label{equ:13}
\end{equation}
Note that $F_{RA_i}$ and $F_{SR_i}$ will be sent to the next RAB and SRB, respectively. We have two output projectors $OP_r$ and $OP_s$ in RIE, which correspond to RAB and SRB, respectively. The $OP_r$ consists of an FC layer whose size is $K{\times} K{\times} M{\times}(n_r{+}n_c)$, where $K{\times} K$ refers to the number of divided grids of an input character image, $M$ represents the number of anchor boxes in each grid, $n_r$ is the number of radical categories in the datasets, and $n_c$ records the coordinates of the radical location ($x, y, w, h$) and the confidence of radical detection, thus $n_c {=} 5$. 
$OP_s$ consists of two convolutional layers and an FC layer to further handle features with mixed character semantics.

\noindent\textbf{Loss Functions.}
RIE contains two output projectors $OP_r$ and $OP_s$. Thus, we have corresponding loss functions $\mathcal{L}_{R}$ and $\mathcal{L}_{S}$, respectively. $\mathcal{L}_{R}$ consists of three components, $\mathcal{L}_{r}$, $\mathcal{L}_{coo}$ and $\mathcal{L}_{isR}$. We define $\mathcal{L}_{r}$ as the cross-entropy loss 
% \cite{de2005tutorial} 
of radical classification:
\begin{equation}
\small
    \!\mathcal{L}_{r} \!=  \!\sum_{i=0}^{K^2} \sum_{j=0}^{M} \mathbb{I}_{i,j}^{isR} \sum_{r\in R} [p_i(r)log(\hat{p_i}(r)) +  \!(1-p_i(r))log(1-\hat{p_i}(r))],
\label{equ:2}
\end{equation}
where $r$ is a category from radical set $R$; $\mathbb{I}_{i,j}^{isR}$ is assigned 1 or 0 to indicate whether the target radical exists in the proposal coordinates; $p_i(r)$ and $\hat{p_i}(r)$ refer to the probability of target and predicted results, respectively. We also define $\mathcal{L}_{coo}$ as the loss function of the radical coordinates:
\begin{equation}
\begin{aligned}
\small
    \mathcal{L}_{coo} = & \sum_{i=0}^{K^2} \sum_{j=0}^{M} \mathbb{I}_{i,j}^{isR} (2-w_i h_i)[(x_i - \hat{x_i})^2 + (y_i - \hat{y_i})^2] \\
    + & \sum_{i=0}^{K^2} \sum_{j=0}^{M} \mathbb{I}_{i,j}^{isR} (2-w_i h_i) [(w_i - \hat{w_i})^2 +(h_i - \hat{h_i})^2],
\label{equ:3}
\end{aligned}
\end{equation}
$\mathcal{L}_{isR}$ is the loss function of the radical detection confidence:
\begin{equation}
\small
    \mathcal{L}_{isR} = \sum_{i=0}^{K^2} \sum_{j=0}^{M} \mathbb{I}_{i,j}^{isR} (c_i - \hat{c_i})^2 + \lambda \sum_{i=0}^{K^2} \sum_{j=0}^{M} \mathbb{I}_{i,j}^{noR} (c_i - \hat{c_i})^2, \\
\label{equ:4}
\end{equation}
where we consider both cases of radical existence and absence. Thus, we also set $\mathbb{I}_{i,j}^{noR}$ to present if there is no radical covered by proposal coordinates; $\hat{c_i}$ refers to the confidence of radical prediction, and $c_i$ is the existence of radical; $\lambda$ is the weight of absence cases, and we set $\lambda{=}0.05$ in the experiments. Then we define $\mathcal{L}_{S}$ as the loss function of $OP_s$ to learn from categories of structural relations:
\begin{equation}
    \mathcal{L}_{S} = \frac{\lambda_{s}}{N} \sum_{i=0}^{N} q_{i}(s)\log(\hat{q_i}(s)),
\label{equ:5}
\end{equation}
where $N$ is the number of categories, $q_i(s)$ is the probability of target structural relation, while $\hat{q_i}(s)$ is the predicted one.
To sum up, the full loss function of RIE can be represented as: 
$\mathcal{L}_{RIE} {=} \mathcal{L}_{r} {+} \mathcal{L}_{coo} {+} \mathcal{L}_{isR} {+} \mathcal{L}_{S}$.

\subsection{Knowledge Graph Reasoner}
Inspired by graph-based recommendation methods \cite{wang2020mrp2rec,shi2020learning}, we propose KGR, which aims to achieve zero-shot character recognition via a reasoning-based strategy rather than hard-matching strategies. A weight fusion-based reasoning algorithm is proposed to obtain better recognition performance, where a character knowledge graph (CKG) is exploited to organize character information. We also utilize the predictions from RIE as soft labels to improve the adaptive ability of the reasoning process.

\noindent\textbf{Character Knowledge Graph.} We intend to reuse existing knowledge graphs of characters in various languages, including Oracle, Bronze \cite{chi2022zinet}, Korean, and simplified Chinese\footnote{http://humanum.arts.cuhk.edu.hk/Lexis/lexi-mf/}, where we can extract the radical composition of characters and structural relations. To achieve reasoning-based character recognition, we focus on three kinds of entities in these large-scale CKGs, i.e., character, radical, and structure\footnote{CKGs record the structure of a character, which is considered equal to our defined structural relation.}.
Note that characters are associated with the corresponding radicals by the relation \textit{contain}, and the structures are connected with characters by \textit{compose}, which provides the possibility of more flexible reasoning via CKG.

\noindent\textbf{Weight Fusion-based Reasoning.}
The inputs of the weight fusion-based reasoning algorithm $CharReason(\cdot)$ are $CKG$, radical predictions ($RPs$), and structural relation prediction ($SP$). The predictions of radical categories $RPs {\in} \mathbb{R}^{num{\times}n_r}$ are extracted from the output of the projector $OP_r$, where $num$ is the number of radicals recognized by RIE and $n_r$ is the length of each prediction. $SP$ is the output of the projector $OP_r$ with the length $n_s$, which is the number of predefined structural relations.

As shown in the Algorithm \ref{alg:1}, firstly, we map the predictions in each $RP$ to generate the candidate radical mappings $M$, where the confidence of each $m_i {\in} M$ is calculated by:
\begin{equation}
    m_i.conf= \frac{1}{num}\sum_{i=1}^{num} p_r,
\label{equ:8}
\end{equation}
where $p_r$ is the prediction confidence $conf$ of each candidate radical in the mapping $m_i$. We search for candidate characters $C_r$ that match the radicals in $m_i$ via $searchRad(\cdot)$ and candidate characters $C_s$ that match the relations $sp_j$ via $searchStr(\cdot)$ in CKG, where $sp_j {\in} SP$. Note that we sort $M$ and $SP$ via $maxSort(\cdot)$ based on the value of prediction confidence before searching to speed up the reasoning process. As a result, we have candidate character $C_t$ that satisfies both radical mapping and structural relation. Its confidence $p_c$ is calculated by a weighted fusion of the $m_i$ and $sp_j$ in line 7, where $\theta{=}0.7$. The obtained $C_t$ and the corresponding confidences $p_c$ are stored in the character prediction $PC$. The algorithm outputs the sorted $PC$ as the final recognition result, where we have the confidence of all candidates $C_t$ to maximize the possibility of correct recognition.

Our proposed KGR comprehensively considers the extracted character information from RIE, which effectively alleviates the low-precision character reasoning issue caused by hard-matching strategies. Note that KGR supports adding new character categories by updating CKG without additional model training. Thus, we consider RZCR as a zero-shot method that is able to recognize unseen character categories by exploiting CKG with radical information and reasoning-based strategy.

\begin{algorithm}[!t] \small
\caption{Weight Fusion-based Reasoning. \\
$PC = CharReason(CKG, RPs, SP)$}
\label{alg:1}

\begin{algorithmic}[1] %[1] enables line numbers

\REQUIRE ~~\\ 
        Character knowledge graph $CKG$;
        Radical predictions $RPs$;
        Structural relations prediction $SP$.\\
\ENSURE ~~\\ 
Character predictions with confidence $PC$.\\

\STATE $M = map(RPs)$; 
% \{candidate radical mappings $M$ are generated from $RPs$.\}
\FOR{each $m_i \in maxSort(M)$}
\STATE $C_r = searchRad(m_i, CKG)$; 
% \{Search for candidate characters $C_r$ that match the mappings $m_i$ in CKG.\}
\FOR{each $sp_j \in maxSort(SP)$}
\STATE $C_s =  searchStr (sp_j, CKG)$; 
% \{Search for candidate characters $C_s$ that match the structural relation $p_j$.\}
\STATE  $C_t = C_r \cap C_s $ \\
\STATE $p_{c} = \theta m_i.conf +  (1 - \theta)sp_j.conf$; \\
\STATE $PC.add(C_t, p_{c})$  \\
\ENDFOR
\ENDFOR
\RETURN $maxSort(PC)$ \\
\end{algorithmic}
\end{algorithm}
\vspace{-0.3cm}

        % threshold $\varepsilon_{min}$;
% \STATE $rel_{ct},rel_{cp}$= {\rm getRelation}{($CKG$)};

% \STATE    \qquad  $p_c = \frac{\theta}{m} \sum_{i=1}^m p_j^i +  (1 - \theta)q_k$; \\

% \IF {\rm getQuery ($Radical set, rel_{ct}$)}
% \STATE $C, p_i$ = {\rm getCharacter}{($r_{comb}, rel_{ct}$)}; \\

% \STATE $C$ = {\rm getCharacterFilter}($C, q_j, rel_{cp}$); \\
% \STATE $set_r$ = Getcontent($p_i$) \{$set_r$ is the Radical combination set represented by $p_i$\}
% \STATE s = Getcontent($q_j$) \{S is the structural relationship represented by $q_j$\}

% \IF {\rm getCharacter($p_i, rel_{ct}, q_j, rel_{cp}$)} %\{Retrieve matching characters in CKG\}
% \STATE  $p_c = p_t$;\\
% \ELSE
% \IF {$p_t \textgreater \varepsilon_{min}$}
% \STATE  $n_{rc}$ = {\rm getQuery}{($p_i, rel_{ct}$)};       \\
% \STATE $p_c = \frac{\alpha}{n_{rc}} p_t$; \\
% \ELSE
% \STATE $p_c = \beta p_t$; \\
% \ENDIF
% \ENDIF

\section{Experiments}
\label{sec:Experiments}

\begin{table*}[!t]
\centering
	\resizebox{1\linewidth}{!}{% 
\begin{tabular}{@{}lcccccccccc@{}}
\toprule

\multirow{2}{*}{Method}  & \multicolumn{4}{c}{OracleRC}  & \multicolumn{2}{c}{ICDAR2013}    & \multicolumn{2}{c}{CTW}    & \multicolumn{2}{c}{HWDB1.1} \\

\cmidrule(l){2-5} \cmidrule(l){6-7}  \cmidrule(l){8-9}  \cmidrule(l){10-11} 
 & Top-1 & Top-3 & Top-5 & Cat$_{Avg}$  &  Top-1 &  Cat$_{Avg}$  & Top-1  & Cat$_{Avg}$     & Top-1  & Cat$_{Avg}$     \\ \midrule

% HOG+SVM \cite{dalal2005histograms}   & 9.29\%   & 12.75\%  & 24.35\%     &  1\%   & 2\%     & 3\%    & 04\%     & 05\%   &  06\%  &  07\%  \\
AlexNet \cite{krizhevsky2012imagenet} & 26.93\% & 36.45\% & 40.03\%  & 21.74\% & 89.99\% & 80.14\% & 76.49\%  & 61.28\%  & 88.74\%  & 85.32\% \\
VGG16 \cite{simonyan2014very}        & 27.75\% & 38.12\% & 41.53\% &  20.38\% & 90.68\%  & 82.76\%  & 79.38\%  & 68.34\% & 89.67\%  &  84.60\% \\
HCCR-GoogLeNet \cite{zhong2015high}  & 28.52\% & 36.75\% & 39.86\%    &  18.81\%  & 96.26\% & 88.97\% & 82.28\%  & 71.21\% &  94.85\% &  90.36\%  \\
DropSample-DCNN\cite{yang2016dropsample} & 29.19\% & 39.27\%  & 42.03\%  & 19.59\%  & 97.23\% &  89.10\% & 82.37\% & 71.21\% & {\color[HTML]{FF0000}96.57\%} & 91.42\% \\
ResNet \cite{he2016deep}             & 28.50\% & 33.02\%  & 40.66\%  &  21.98\%   & 92.18\% &  85.68\%  & 79.46\% & 69.82\% & 90.98\% & 86.06\% \\
% Inception-v4 \cite{szegedy2017inception} & 30.31\% & 49.96\% & 52.13\%   &    & 95.79\%      &           &  80.50\%      &   &  94.21\%    &   \\ 
DenseNet \cite{huang2017densely}     & 27.85\% & 38.63\% & 47.48\% & 19.20\% &  95.90\%  & 90.36\% & 79.88\% & 68.48\% &  94.32\%  &  89.72\%   \\ 
DirectMap\cite{zhang2017online}      & 30.48\% & {\color[HTML]{0070C0}44.89\%} & {\color[HTML]{0070C0}54.72\%} & 23.59\% & {\color[HTML]{0070C0}97.37\%} & 90.62\% & 84.23\% & 72.50\% & 96.25\%   &  91.28\% \\   
M-RBC + IR\cite{yang2017improving}   & 30.53\% & 42.74\% & 49.32\% & 20.72\% & {\color[HTML]{0070C0}97.37\%} & 88.70\% & 83.65\%  & 73.07\% & 96.14\%   &  90.76\% \\ 
\midrule

RAN \cite{zhang2018radical}          & 35.37\% & - & -  & 32.48\% &  93.79\% & 88.69\% & 81.80\%  & 76.59\% &  92.28\% &  89.43\% \\
DenseRAN \cite{wang2018denseran}     & 36.02\% & - & -  & 32.16\% &  96.66\% & {\color[HTML]{0070C0}91.02\%} &  85.56\% & {\color[HTML]{0070C0}82.47\%} &  95.32\% &  {\color[HTML]{0070C0}91.76\%}  \\                     
FewshotRAN \cite{wang2019radical}    & 33.31\% & - & -  & 30.90\% &  96.97\% & 90.42\% &  86.78\% & 81.16\% & 96.32\% &  91.59\%  \\              
HDE-Net \cite{cao2020zero}           & {\color[HTML]{0070C0}36.79\%} & - & - & {\color[HTML]{0070C0}33.10\%} &  96.74\%  & 88.75\% &  {\color[HTML]{FF0000}89.25\%}  & 78.94\% &  95.63\% &  90.48\%  \\                    
Stroke-to-Character \cite{chen2021zero} & 27.30\% & - & - & 20.09\% &  96.28\% & 89.11\% &  85.29\% & 77.14\% & 93.97\% & 89.83\% \\ 
STAR \cite{zeng2022star} & 30.65\% & - & - & 23.76\% &  97.11\% & 88.82\% &  85.43\% & 78.26\% & 94.24\% & 90.03\% \\ 
HCRN\cite{huang2022hippocampus} & 35.84\% & - & - & 32.24\% &  96.70\% & 88.97\% &  85.59\% & 77.62\% & 95.86\% & 91.67\% \\ \midrule

RZCR (Ours) & {\color[HTML]{FF0000}61.36\%} & {\color[HTML]{FF0000}71.02\%} & {\color[HTML]{FF0000}74.39\%}  & {\color[HTML]{FF0000}58.84\%} &  {\color[HTML]{FF0000}97.42\%}  & {\color[HTML]{FF0000}91.43\%} & {\color[HTML]{0070C0}88.74\%}  & {\color[HTML]{FF0000}82.89\%} &  {\color[HTML]{0070C0}96.45\%} &  {\color[HTML]{FF0000}92.28\%}   \\ \bottomrule
\end{tabular}}
\caption{Quantitative comparisons with state-of-the-art methods on four datasets. The best and second-best results are highlighted in {\color[HTML]{FF0000} red} and {\color[HTML]{0070C0} blue} colors, respectively.}
\label{tab:1}
\end{table*}

\begin{table*}[!t]
\centering
\resizebox{0.9\linewidth}{!}{
\begin{tabular}{@{}lccccccccc@{}}
\toprule
\multirow{2}{*}{Method}  & \multicolumn{3}{c}{OracleRC} & \multicolumn{3}{c}{Combined dataset} & \multicolumn{3}{c}{CTW} \\
\cmidrule(l){2-4} \cmidrule(l){5-7} \cmidrule(l){8-10} & c-500 & c-1000 & c-1205 & c-1000 & c-2000 & c-2755 & c-1000 & c-2000 & c-3150  \\ \midrule
DenseRAN \cite{wang2018denseran} & 5.28\% & 10.67\% & 11.58\% & 8.44\% & 19.51\% & 30.68\% & 0.54\% & 1.95\% & 5.39\%           \\
HDE-Net \cite{cao2020zero}       & 7.12\% & 9.76\% & 10.51\% & 12.77\% & 25.13\% & 33.49\%  & 2.11\% & 6.96\% & 7.75\%           \\
Stroke-to-Character \cite{chen2021zero}  & 3.37\% & 7.48\% & 7.79\% & 13.85\% & 25.73\% & 37.91\% & 2.54\% & 6.82\% & 8.61\%          \\
STAR \cite{zeng2022star} & 6.14\% & 10.62\% & 12.23\% & 19.47\% & 35.53\% & 43.86\% & 3.77\% & 11.00\% & 11.27\%          \\
RZCR (Ours)     & \textbf{39.21\%} & \textbf{52.43\%} & \textbf{54.28\%} & \textbf{65.58\%} & \textbf{73.56\%} & \textbf{78.75\%} & \textbf{49.73\%} & \textbf{61.39\%} & \textbf{64.82\%}  \\ \bottomrule
\end{tabular}}
\caption{Comparisons with zero-shot character recognition methods.}
\label{tab:2}
\end{table*}

\subsection{Experimental Setup}
\noindent\textbf{Datasets.}  
To evaluate our method on real-world character image sets which suffer from the few-sample problem, we introduce a new character image dataset called OracleRC. We collect oracle rubbing images from \cite{wu2012} and normalize images by a denoising method \cite{shi2022charformer}. OracleRC includes 2,005 character categories that can be decomposed by 202 radicals and 14 structural relations, where the number of character samples ranges from 1 to 32. Radicals and structural relations were manually annotated by 8 linguists. We also validate RZCR on handwritten Chinese datasets ICDAR2013 \cite{yin2013icdar}, HWDB1.1 \cite{liu2013online}, scene character dataset CTW \cite{yuan2019large}, and Korean dataset PE92 \cite{kim1996handwritten} for a comprehensive evaluation. To increase the adaptability of RZCR, we propose a radical splicing-based synthetic character strategy to enlarge the training set and reduce the cost of human annotations.

\noindent\textbf{Implementation Details.} The resolution of the input image is $416 {\times} 416$. We exploit data enhancement strategies, including translation, rotation, scaling, and background transformation. Parameters $K {=} 13$ and $M {=} 3$ are set in RIE. All experiments are conducted with Adadelta optimization where the hyperparameters are set to $\rho {=} 0.95$ and $\varepsilon {=} 10^{-6}$.

\subsection{Experimental Results}
We compare RZCR with state-of-the-art character recognition methods on four datasets, and the results are shown in Table \ref{tab:1}. We output Top-n prediction by confidence to present the average classification accuracy based on samples. Considering a few categories with more samples are not enough to reflect the overall recognition performance in an unbalanced dataset, we also calculate the average accuracy for each category and then average over all categories, i.e., Cat$_{Avg}$. Note that in Table \ref{tab:1}, general OCR methods are presented in the former rows, and the latter are zero-shot recognition methods. For all four datasets, we select 80\% of the samples in each character category as the training set and the remaining as the test set. Note that categories containing only one sample are not included in this experiment since general recognition methods are not able to train on these categories. 

For the few-sample dataset OracleRC, we can find RZCR significantly outperforms all general OCR methods and zero-shot methods on Top-n accuracy and Cat$_{Avg}$. We also find zero-shot methods except Stroke-to-Character gain higher accuracy than general methods on Top-1 accuracy and Cat$_{Avg}$. Note that lower Cat$_{Avg}$ obtained by general OCR methods means that these methods perform poorly on categories with few samples. The main reason for this disparity is that an insufficient amount of training data limits their performance. In contrast, for zero-shot recognition methods, decomposing characters into elements brings an increasing number of training samples and a decrease in training categories, which alleviates the few-sample issue. Meanwhile, the better performance of RZCR among zero-shot methods benefits from the powerful KGR with a flexible reasoning strategy, as we discussed before. We also evaluate RZCR on ICDAR2013, CTW, and HWDB1.1 to demonstrate its adaptability, where the same training and testing setup are conducted for all methods. As shown in Table \ref{tab:1}, both general and zero-shot methods are competitive in these cases with sufficient training samples, where our RZCR also gains promising results. We find RZCR performs better than others on metric Cat$_{Avg}$, which proves our method is less influenced by categories with different numbers of samples during recognition. 
% Besides, we compare the Top-1 accuracy of rare categories (last 30\%) with other methods. In CTW, RZCR (78.35\%) achieves a significant improvement compared to the best-performing method HDE-Net (64.56\%), which demonstrates RZCR is effective for handling tail data.

Then, we conduct an additional experiment to demonstrate the validity of RZCR on zero-shot character recognition, and the results are presented in Table \ref{tab:2}. Only zero-shot recognition methods are included in this experiment since general methods are not able to recognize unseen character categories. Three datasets are applied, including OracleRC, CTW, and a combined dataset by two handwritten character datasets ICDAR2013 and HWDB1.1. We follow the experimental setting introduced by previous zero-shot methods \cite{wang2018denseran,chen2021zero}, 
% and gradually increase the number of training categories. 
where we fix the test categories at the beginning and gradually increase the number of training categories. 
Thus, in this experiment, 800 out of 2,005 categories in OracleRC are unseen categories used for testing. Similarly, 1,000/3,755 and 500/3,650 categories are applied for testing on the combined dataset and CTW, respectively. Meanwhile, as shown in Table \ref{tab:2}, c-$m$ refers to the comparing group which applied $m$ training categories, $m {\in} \{500, 1000, ...\}$. We can find that our proposed RZCR significantly surpasses other sequence-based zero-shot methods in each set of datasets, which further shows the superiority of our reasoning-based recognition strategy. More specifically, RZCR benefits from the CKG-based knowledge organization and a flexible reasoning algorithm, which results in the possibility of correct character recognition even with incorrect Top-1 radical prediction. Thus, we can conclude that our proposed RZCR, as a generic character recognition method, is effective for different character datasets, particularly for few-sample datasets.

\subsection{Ablation Study}

\begin{table}[!t]
\centering
	\resizebox{0.95\linewidth}{!}{% 
\begin{tabular}{@{}lcccc@{}}
\toprule
  & Top-1 RPs   &    Top-1 RPs \& SP     & RPs    &  RPs \& SP (Ours) \\ \midrule
$Acc_C$        & 38.56\%          & 37.02\%        & 58.06\%     & \textbf{61.36}\%  \\ \bottomrule
\end{tabular}}
\caption{Ablation study on KGR.}
\label{tab:4}
\end{table}

\begin{table}[!t]
\centering
	\resizebox{0.70\linewidth}{!}{% 
\begin{tabular}{@{}ccccccc@{}}
\toprule
  & Radicals             & Characters  & Advances    \\ \midrule
ICDAR      & 95.94\%          & 97.42\%      & + \textbf{1.48\%}  \\
CTW        & 84.86\%          & 88.74\%      & + \textbf{3.88\%}   \\
OracleRC   & 58.59\%          & 61.36\%      & + \textbf{2.47\%}   \\ \bottomrule
\end{tabular}}
\caption{Comparison of radical and character recognition, to prove the effect of KGR.}
\label{tab:rea-1}
\end{table}

\begin{table}[!t]\footnotesize
\centering
	\resizebox{0.70\linewidth}{!}{% 
\begin{tabular}{@{}lcc@{}}
\toprule
  & Without FAL & With FAL (Ours) \\ \midrule
$Acc_R$/$Acc_{SR}$ & 55.87\%/70.64\% &  \textbf{58.59\%}/\textbf{79.41\%} \\ \bottomrule
\end{tabular}}
\caption{Comparisons on baseline networks in RIE.}
\label{tab:c}
\end{table}

\noindent\textbf{Impact of Reasoning Strategies in KGR.}
We also conduct experiments to validate the effectiveness of reasoning-based character recognition. Firstly, we change the radical information and matching strategy in KGR. In Table~\ref{tab:4}, ``Top-1 RPs'' refers to a hard-matching strategy that considers only radicals with the highest confidence in RPs, while ``Top-1 RPs \& SP'' utilizes both radicals and structural relations with the highest confidence in RPs and SP, respectively. ``RPs'' considers all candidate radicals in RPs for reasoning, while ``RPs \& SP'' exploits all candidates in both RPs and SP, i.e., our reasoning-based algorithm. We see that the character recognition accuracy $Acc_C$ of the latter two is significantly higher than that of the former two, which shows our reasoning-based algorithm surpasses hard-matching strategies. We can also find the increasing number of elements limits the performance of the hard-matching strategy, while the accuracy of our algorithm increases when adding structural relations. A similar phenomenon can also be found in Table \ref{tab:rea-1}, where character recognition results by RZCR are higher than radical recognition results only using RIE since the reasoning-based KGR can maximize the possibility of correct recognition. 

\noindent\textbf{Impact of Baseline Networks in RIE.} 
RP and SP are output by $OP_r$ and $OP_s$ via RABs and SRBs, respectively, in RIE, where the two features are interacted by the FAL layer. As shown in Table \ref{tab:c}, we apply an ablation test to present the superiority of the proposed FAL. ``Without FAL'' outputs RPs and SP by two baseline networks, respectively. ``With FAL" indicates that RABs and SRBs exchange semantic information by FAL, i.e., RIE.  $Acc_R$ and $Acc_{SR}$ refer to the recognition accuracy of the radical categories and structural relations. We find that RIE obtains better performance in both metrics. This indicates that the feature extraction of radicals and structures can be improved mutually since the structural relation is relevant to the radical location. 

\begin{figure}[!t]
	\centering
	\includegraphics[width=0.90\linewidth]{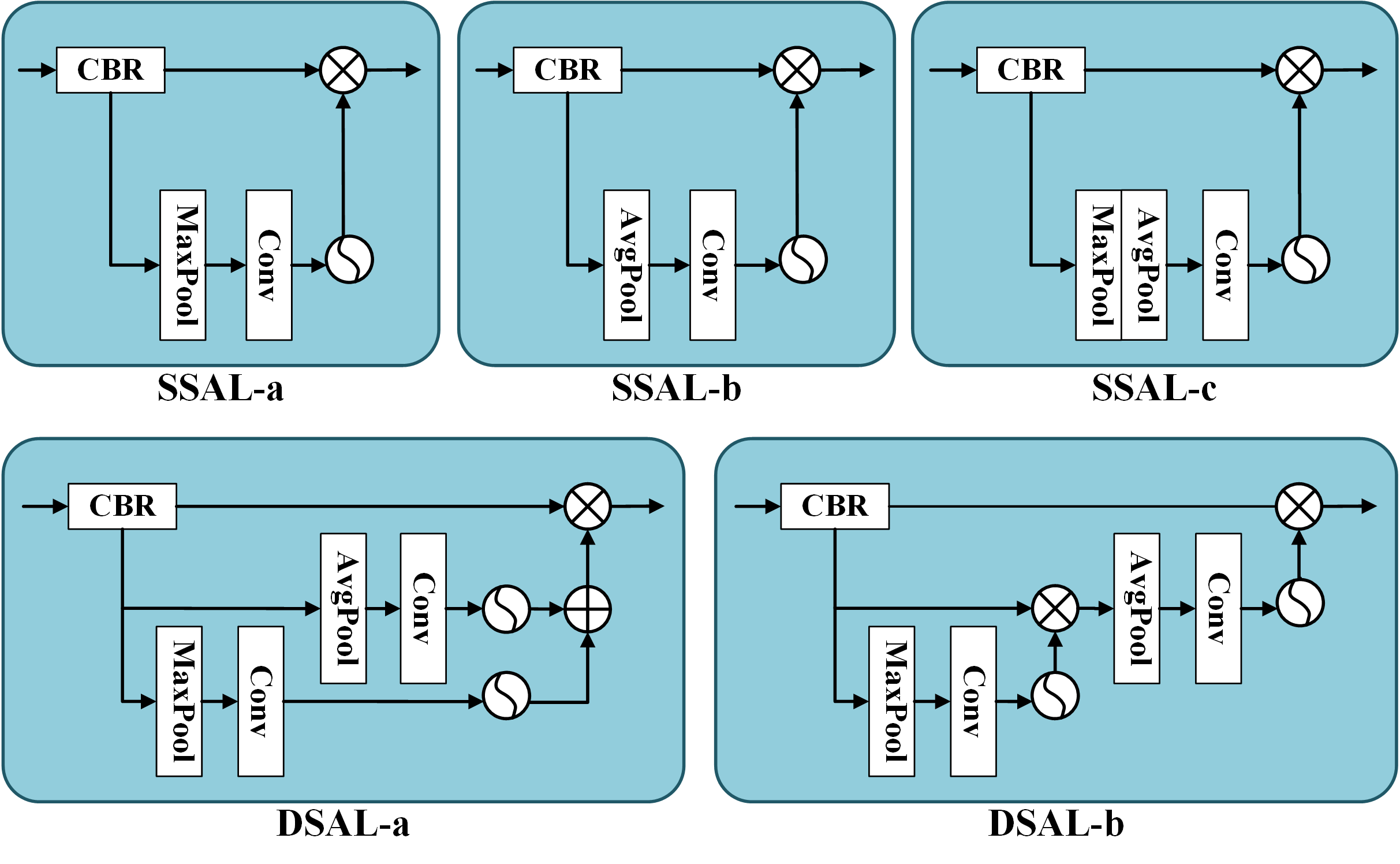}
    \caption{Five connection schemes for the attention layer.
    \label{fig:4}}
\end{figure}

\begin{table}[!t]
  \centering
        \resizebox{0.9\linewidth}{!}{
        \begin{tabular}{@{}lccccc@{}}
        \toprule
         & SSAL-a & SSAL-b & SSAL-c & DSAL-a & DSAL-b (Ours)
            \\ \midrule
        $Acc_R$ & 56.94\% & 57.18\% & 57.83\% & 58.36\% & \textbf{58.59}\% \\ \bottomrule
        \end{tabular}}
        \makeatletter\def\@captype{table}\makeatother\caption{Ablation study on attention layer schemes.}
        \label{tab:5}
\end{table}

\noindent\textbf{Impact of Attention Layer Schemes.}
In RIE, we aim to utilize the attention mechanism to refine the radical information extraction on character images with overlapping radicals and unclear boundaries. As shown in Fig.~\ref{fig:4}, we design five possible schemes to stack attention layers, where three schemes are proposed based on the single spatial attention layer (SSAL) and the remaining two are based on the DSAL. The results of the radical recognition by different schemes are shown in Table~\ref{tab:5}. We conduct this experiment on the OracleRC dataset, where we record the radical recognition accuracy $Acc_R$. It is easy to find that DSAL-based schemes outperform three SSAL-based schemes. That is because DSALs consider both foreground information for radical feature selection and background information to maintain the integrity of radicals. Thus, we select better-performed DSAL-b as the attention layer in RABs.

% add in camera ready
% According to the ablation experimental results, we find that the soft reasoning KGR component (24.3\% improvement) and attention layer FAL (8.7\% improvement) are the two largest contributors to the effectiveness of RZCR. Besides, the setting of two output projectors in RIE and the attention layer DSAL applied in RABs also improves the performance of RZCR to some extent.

\section{Conclusion}
\label{sec:Conclusion}
In this paper, we first introduce the importance of radical information for character recognition and discuss two character decomposition strategies. Then, we propose RZCR, a novel zero-shot character recognition method, to deal with unbalanced character datasets with few samples. RZCR obtains promising experimental results, especially on few-sample datasets. In the future, we will engage to generalize RZCR to other areas where data can be decomposed into components.

\clearpage

% \appendix

% \section*{Ethical Statement}

% There are no ethical issues.

\section*{Acknowledgments}
This work is supported by the National Natural Science Foundation of China (No.62077027), the Department of Science and Technology of Jilin Province, China (20230201086GX), the EU’s Horizon 2020 FET proactive project (grant agreement No.823783), and the ``Paleography and Chinese Civilization Inheritance and Development Program" Collaborative Innovation Platform (No.G3829). This work is also supported by Professor Chuntao Li and his team from the School of Archeology, Jilin University.

%% The file named.bst is a bibliography style file for BibTeX 0.99c
\bibliographystyle{named}
\bibliography{ijcai23}

\begin{thebibliography}{}

\bibitem[\protect\citeauthoryear{Akata \bgroup \em et al.\egroup
  }{2015}]{akata2015label}
Zeynep Akata, Florent Perronnin, Zaid Harchaoui, and Cordelia Schmid.
\newblock Label-embedding for image classification.
\newblock {\em IEEE TPAMI}, 38(7):1425--1438, 2015.

\bibitem[\protect\citeauthoryear{Anderson}{2006}]{anderson2006long}
Chris Anderson.
\newblock {\em The long tail: Why the future of business is selling less of
  more}.
\newblock Hachette Books, 2006.

\bibitem[\protect\citeauthoryear{Balaha \bgroup \em et al.\egroup
  }{2021}]{balaha2021new}
Hossam~Magdy Balaha, Hesham~Arafat Ali, Mohamed Saraya, and Mahmoud Badawy.
\newblock A new arabic handwritten character recognition deep learning system
  (ahcr-dls).
\newblock {\em Neural Computing and Applications}, 33(11):6325--6367, 2021.

\bibitem[\protect\citeauthoryear{Cao \bgroup \em et al.\egroup
  }{2020}]{cao2020zero}
Zhong Cao, Jiang Lu, Sen Cui, and Changshui Zhang.
\newblock Zero-shot handwritten chinese character recognition with hierarchical
  decomposition embedding.
\newblock {\em Elsevier PR}, 107:107488, 2020.

\bibitem[\protect\citeauthoryear{Chen \bgroup \em et al.\egroup
  }{2021}]{chen2021zero}
Jingye Chen, Bin Li, and Xiangyang Xue.
\newblock Zero-shot chinese character recognition with stroke-level
  decomposition.
\newblock In {\em IJCAI}, 2021.

\bibitem[\protect\citeauthoryear{Chi \bgroup \em et al.\egroup
  }{2022}]{chi2022zinet}
Yang Chi, Fausto Giunchiglia, Daqian Shi, Xiaolei Diao, Chuntao Li, and Hao Xu.
\newblock Zinet: Linking chinese characters spanning three thousand years.
\newblock In {\em ACL}, pages 3061--3070, 2022.

\bibitem[\protect\citeauthoryear{Cire{\c{s}}an and
  Meier}{2015}]{cirecsan2015multi}
Dan Cire{\c{s}}an and Ueli Meier.
\newblock Multi-column deep neural networks for offline handwritten chinese
  character classification.
\newblock In {\em IJCNN}, 2015.

\bibitem[\protect\citeauthoryear{G{\'e}ron}{2019}]{geron2019hands}
Aur{\'e}lien G{\'e}ron.
\newblock {\em Hands-on machine learning with Scikit-Learn, Keras, and
  TensorFlow: Concepts, tools, and techniques to build intelligent systems}.
\newblock O'Reilly Media, 2019.

\bibitem[\protect\citeauthoryear{He \bgroup \em et al.\egroup
  }{2016}]{he2016deep}
Kaiming He, Xiangyu Zhang, Shaoqing Ren, and Jian Sun.
\newblock Deep residual learning for image recognition.
\newblock In {\em CVPR}, 2016.

\bibitem[\protect\citeauthoryear{Ho \bgroup \em et al.\egroup
  }{2003}]{ho2003radical}
Connie Suk-Han Ho, Ting-Ting Ng, and Wing-Kin Ng.
\newblock A “radical” approach to reading development in chinese: The role
  of semantic radicals and phonetic radicals.
\newblock {\em Journal of literacy research}, 35(3):849--878, 2003.

\bibitem[\protect\citeauthoryear{Huang \bgroup \em et al.\egroup
  }{2017}]{huang2017densely}
Gao Huang, Zhuang Liu, Laurens Van Der~Maaten, and Kilian~Q Weinberger.
\newblock Densely connected convolutional networks.
\newblock In {\em CVPR}, 2017.

\bibitem[\protect\citeauthoryear{Huang \bgroup \em et al.\egroup
  }{2019}]{huang2019obc306}
Shuangping Huang, Haobin Wang, Yongge Liu, Xiaosong Shi, and Lianwen Jin.
\newblock Obc306: A large-scale oracle bone character recognition dataset.
\newblock In {\em ICDAR}, 2019.

\bibitem[\protect\citeauthoryear{Huang \bgroup \em et al.\egroup
  }{2022}]{huang2022hippocampus}
Guanjie Huang, Xiangyu Luo, Shaowei Wang, Tianlong Gu, and Kaile Su.
\newblock Hippocampus-heuristic character recognition network for zero-shot
  learning in chinese character recognition.
\newblock {\em Pattern Recognition}, page 108818, 2022.

\bibitem[\protect\citeauthoryear{KIM \bgroup \em et al.\egroup
  }{1996}]{kim1996handwritten}
Dae-Hwan KIM, Young-Sup Hwang, Sang-Tae Park, Eun-Jung Kim, Sang-Hoon Paek, and
  Sung-Yang BANG.
\newblock Handwritten korean character image database pe92.
\newblock {\em IEICE TOIS}, 79(7):943--950, 1996.

\bibitem[\protect\citeauthoryear{Krizhevsky \bgroup \em et al.\egroup
  }{2012}]{krizhevsky2012imagenet}
Alex Krizhevsky, Ilya Sutskever, and Geoffrey~E Hinton.
\newblock Imagenet classification with deep convolutional neural networks.
\newblock In {\em NeurIPS}, 2012.

\bibitem[\protect\citeauthoryear{Lampert \bgroup \em et al.\egroup
  }{2009}]{lampert2009learning}
Christoph~H Lampert, Hannes Nickisch, and Stefan Harmeling.
\newblock Learning to detect unseen object classes by between-class attribute
  transfer.
\newblock In {\em CVPR}, 2009.

\bibitem[\protect\citeauthoryear{Lampert \bgroup \em et al.\egroup
  }{2013}]{lampert2013attribute}
Christoph~H Lampert, Hannes Nickisch, and Stefan Harmeling.
\newblock Attribute-based classification for zero-shot visual object
  categorization.
\newblock {\em IEEE TPAMI}, 36(3):453--465, 2013.

\bibitem[\protect\citeauthoryear{Liu \bgroup \em et al.\egroup
  }{2011}]{liu2011casia}
Cheng-Lin Liu, Fei Yin, Da-Han Wang, and Qiu-Feng Wang.
\newblock Casia online and offline chinese handwriting databases.
\newblock In {\em ICDAR}, 2011.

\bibitem[\protect\citeauthoryear{Liu \bgroup \em et al.\egroup
  }{2013}]{liu2013online}
Cheng-Lin Liu, Fei Yin, Da-Han Wang, and Qiu-Feng Wang.
\newblock Online and offline handwritten chinese character recognition:
  benchmarking on new databases.
\newblock {\em Elsevier PR}, 46(1):155--162, 2013.

\bibitem[\protect\citeauthoryear{Luo \bgroup \em et al.\egroup
  }{2020}]{Luo_2020_CVPR}
Canjie Luo, Yuanzhi Zhu, Lianwen Jin, and Yongpan Wang.
\newblock Learn to augment: Joint data augmentation and network optimization
  for text recognition.
\newblock In {\em CVPR}, June 2020.

\bibitem[\protect\citeauthoryear{Lyu \bgroup \em et al.\egroup
  }{2017}]{lyu2017auto}
Pengyuan Lyu, Xiang Bai, Cong Yao, Zhen Zhu, Tengteng Huang, and Wenyu Liu.
\newblock Auto-encoder guided gan for chinese calligraphy synthesis.
\newblock In {\em ICDAR}, 2017.

\bibitem[\protect\citeauthoryear{Mushtaq \bgroup \em et al.\egroup
  }{2021}]{mushtaq2021urdudeepnet}
Faisel Mushtaq, Muzafar~Mehraj Misgar, Munish Kumar, and Surinder~Singh
  Khurana.
\newblock Urdudeepnet: offline handwritten urdu character recognition using
  deep neural network.
\newblock {\em Neural Computing and Applications}, 33(22):15229--15252, 2021.

\bibitem[\protect\citeauthoryear{Myers}{2016}]{myers2016knowing}
James Myers.
\newblock Knowing chinese character grammar.
\newblock {\em Elsevier Cognition}, 147:127--132, 2016.

\bibitem[\protect\citeauthoryear{Qu \bgroup \em et al.\egroup
  }{2018}]{qu2018data}
Xiwen Qu, Weiqiang Wang, Ke~Lu, and Jianshe Zhou.
\newblock Data augmentation and directional feature maps extraction for in-air
  handwritten chinese character recognition based on convolutional neural
  network.
\newblock {\em Pattern recognition letters}, 111:9--15, 2018.

\bibitem[\protect\citeauthoryear{Rohrbach \bgroup \em et al.\egroup
  }{2011}]{rohrbach2011evaluating}
Marcus Rohrbach, Michael Stark, and Bernt Schiele.
\newblock Evaluating knowledge transfer and zero-shot learning in a large-scale
  setting.
\newblock In {\em CVPR}, 2011.

\bibitem[\protect\citeauthoryear{Shanthi and
  Duraiswamy}{2010}]{shanthi2010novel}
N~Shanthi and K~Duraiswamy.
\newblock A novel svm-based handwritten tamil character recognition system.
\newblock {\em Springer Pattern Analysis and Applications}, 13(2):173--180,
  2010.

\bibitem[\protect\citeauthoryear{Shen}{2005}]{shen2005investigation}
Helen~H Shen.
\newblock An investigation of chinese-character learning strategies among
  non-native speakers of chinese.
\newblock {\em Elsevier System}, 33(1):49--68, 2005.

\bibitem[\protect\citeauthoryear{Shi \bgroup \em et al.\egroup
  }{2020}]{shi2020learning}
Daqian Shi, Ting Wang, Hao Xing, and Hao Xu.
\newblock A learning path recommendation model based on a multidimensional
  knowledge graph framework for e-learning.
\newblock {\em Knowledge-Based Systems}, 195:105618, 2020.

\bibitem[\protect\citeauthoryear{Shi \bgroup \em et al.\egroup
  }{2022a}]{shi2022charformer}
Daqian Shi, Xiaolei Diao, Lida Shi, Hao Tang, Yang Chi, Chuntao Li, and Hao Xu.
\newblock Charformer: A glyph fusion based attentive framework for
  high-precision character image denoising.
\newblock In {\em ACM MM}, 2022.

\bibitem[\protect\citeauthoryear{Shi \bgroup \em et al.\egroup
  }{2022b}]{shi2022rcrn}
Daqian Shi, Xiaolei Diao, Hao Tang, Xiaomin Li, Hao Xing, and Hao Xu.
\newblock Rcrn: Real-world character image restoration network via skeleton
  extraction.
\newblock In {\em ACM MM}, 2022.

\bibitem[\protect\citeauthoryear{Simonyan and
  Zisserman}{2015}]{simonyan2014very}
Karen Simonyan and Andrew Zisserman.
\newblock Very deep convolutional networks for large-scale image recognition.
\newblock In {\em ICLR}, 2015.

\bibitem[\protect\citeauthoryear{Su and Wang}{2003}]{su2003novel}
Yih-Ming Su and Jhing-Fa Wang.
\newblock A novel stroke extraction method for chinese characters using gabor
  filters.
\newblock {\em Elsevier PR}, 36(3):635--647, 2003.

\bibitem[\protect\citeauthoryear{Sun \bgroup \em et al.\egroup
  }{2021}]{sun2021research}
Xiaohong Sun, Jinan Gu, and Hongying Sun.
\newblock Research progress of zero-shot learning.
\newblock {\em Applied Intelligence}, 51(6):3600--3614, 2021.

\bibitem[\protect\citeauthoryear{Wang \bgroup \em et al.\egroup
  }{2018a}]{wang2018denseran}
Wenchao Wang, Jianshu Zhang, Jun Du, Zi-Rui Wang, and Yixing Zhu.
\newblock Denseran for offline handwritten chinese character recognition.
\newblock In {\em ICFHR}, 2018.

\bibitem[\protect\citeauthoryear{Wang \bgroup \em et al.\egroup
  }{2018b}]{wang2018zero}
Xiaolong Wang, Yufei Ye, and Abhinav Gupta.
\newblock Zero-shot recognition via semantic embeddings and knowledge graphs.
\newblock In {\em CVPR}, 2018.

\bibitem[\protect\citeauthoryear{Wang \bgroup \em et al.\egroup
  }{2019a}]{wang2019radical}
Tianwei Wang, Zecheng Xie, Zhe Li, Lianwen Jin, and Xiangle Chen.
\newblock Radical aggregation network for few-shot offline handwritten chinese
  character recognition.
\newblock {\em Elsevier PRL}, 125:821--827, 2019.

\bibitem[\protect\citeauthoryear{Wang \bgroup \em et al.\egroup
  }{2019b}]{wang2019survey}
Wei Wang, Vincent~W Zheng, Han Yu, and Chunyan Miao.
\newblock A survey of zero-shot learning: Settings, methods, and applications.
\newblock {\em ACM TIST}, 10(2):1--37, 2019.

\bibitem[\protect\citeauthoryear{Wang \bgroup \em et al.\egroup
  }{2020}]{wang2020mrp2rec}
Ting Wang, Daqian Shi, Zhaodan Wang, Shuai Xu, and Hao Xu.
\newblock Mrp2rec: Exploring multiple-step relation path semantics for
  knowledge graph-based recommendations.
\newblock {\em IEEE Access}, 8:134817--134825, 2020.

\bibitem[\protect\citeauthoryear{Wu}{2012}]{wu2012}
Zhenfeng Wu.
\newblock {\em Shang and Zhou Bronze Inscriptions and Image Integration}.
\newblock Shanghai Classics Publishing House, 2012.

\bibitem[\protect\citeauthoryear{Xie \bgroup \em et al.\egroup
  }{2019}]{xie2019attentive}
Guo-Sen Xie, Li~Liu, Xiaobo Jin, Fan Zhu, Zheng Zhang, Jie Qin, Yazhou Yao, and
  Ling Shao.
\newblock Attentive region embedding network for zero-shot learning.
\newblock In {\em CVPR}, 2019.

\bibitem[\protect\citeauthoryear{Yang \bgroup \em et al.\egroup
  }{2016}]{yang2016dropsample}
Weixin Yang, Lianwen Jin, Dacheng Tao, Zecheng Xie, and Ziyong Feng.
\newblock Dropsample: A new training method to enhance deep convolutional
  neural networks for large-scale unconstrained handwritten chinese character
  recognition.
\newblock {\em Pattern Recognition}, 58:190--203, 2016.

\bibitem[\protect\citeauthoryear{Yang \bgroup \em et al.\egroup
  }{2017}]{yang2017improving}
Xiao Yang, Dafang He, Zihan Zhou, Daniel Kifer, and C~Lee Giles.
\newblock Improving offline handwritten chinese character recognition by
  iterative refinement.
\newblock In {\em ICDAR}, volume~1, pages 5--10. IEEE, 2017.

\bibitem[\protect\citeauthoryear{Yang \bgroup \em et al.\egroup
  }{2019}]{yang2019symmetry}
Mingkun Yang, Yushuo Guan, Minghui Liao, Xin He, Kaigui Bian, Song Bai, Cong
  Yao, and Xiang Bai.
\newblock Symmetry-constrained rectification network for scene text
  recognition.
\newblock In {\em ICCV}, pages 9147--9156, 2019.

\bibitem[\protect\citeauthoryear{Yeung \bgroup \em et al.\egroup
  }{2016}]{yeung2016orthographic}
Pui-sze Yeung, Connie Suk-han Ho, David Wai-ock Chan, and Kevin Kien-hoa Chung.
\newblock Orthographic skills important to chinese literacy development: The
  role of radical representation and orthographic memory of radicals.
\newblock {\em Reading and Writing}, 29(9):1935--1958, 2016.

\bibitem[\protect\citeauthoryear{Yin \bgroup \em et al.\egroup
  }{2013}]{yin2013icdar}
Fei Yin, Qiu-Feng Wang, Xu-Yao Zhang, and Cheng-Lin Liu.
\newblock Icdar 2013 chinese handwriting recognition competition.
\newblock In {\em ICDAR}, 2013.

\bibitem[\protect\citeauthoryear{Yuan \bgroup \em et al.\egroup
  }{2012}]{yuan2012offline}
Aiquan Yuan, Gang Bai, Lijing Jiao, and Yajie Liu.
\newblock Offline handwritten english character recognition based on
  convolutional neural network.
\newblock In {\em DAS}, 2012.

\bibitem[\protect\citeauthoryear{Yuan \bgroup \em et al.\egroup
  }{2019}]{yuan2019large}
Tai-Ling Yuan, Zhe Zhu, Kun Xu, Cheng-Jun Li, Tai-Jiang Mu, and Shi-Min Hu.
\newblock A large chinese text dataset in the wild.
\newblock {\em Springer JCST}, 34(3):509--521, 2019.

\bibitem[\protect\citeauthoryear{Zatsepin \bgroup \em et al.\egroup
  }{2019}]{zatsepin2019fast}
Michael Zatsepin, Yury Vatlin, Iurii Chulinin, and Aleksei Zhuravlev.
\newblock Fast korean syllable recognition with letter-based convolutional
  neural networks.
\newblock In {\em ICDARW}, volume~7, pages 10--13. IEEE, 2019.

\bibitem[\protect\citeauthoryear{Zeng \bgroup \em et al.\egroup
  }{2022}]{zeng2022star}
Jinshan Zeng, Ruiying Xu, Yu~Wu, Hongwei Li, and Jiaxing Lu.
\newblock Star: Zero-shot chinese character recognition with stroke-and
  radical-level decompositions.
\newblock {\em arXiv preprint arXiv:2210.08490}, 2022.

\bibitem[\protect\citeauthoryear{Zhan and Lu}{2019}]{zhan2019esir}
Fangneng Zhan and Shijian Lu.
\newblock Esir: End-to-end scene text recognition via iterative image
  rectification.
\newblock In {\em CVPR}, pages 2059--2068, 2019.

\bibitem[\protect\citeauthoryear{Zhang \bgroup \em et al.\egroup
  }{2017}]{zhang2017online}
Xu-Yao Zhang, Yoshua Bengio, and Cheng-Lin Liu.
\newblock Online and offline handwritten chinese character recognition: A
  comprehensive study and new benchmark.
\newblock {\em Elsevier PR}, 61:348--360, 2017.

\bibitem[\protect\citeauthoryear{Zhang \bgroup \em et al.\egroup
  }{2018}]{zhang2018radical}
Jianshu Zhang, Yixing Zhu, Jun Du, and Lirong Dai.
\newblock Radical analysis network for zero-shot learning in printed chinese
  character recognition.
\newblock In {\em ICME}, 2018.

\bibitem[\protect\citeauthoryear{Zhang \bgroup \em et al.\egroup
  }{2020}]{zhang2020radical}
Jianshu Zhang, Jun Du, and Lirong Dai.
\newblock Radical analysis network for learning hierarchies of chinese
  characters.
\newblock {\em Elsevier PR}, 103:107305, 2020.

\bibitem[\protect\citeauthoryear{Zhong \bgroup \em et al.\egroup
  }{2015}]{zhong2015high}
Zhuoyao Zhong, Lianwen Jin, and Zecheng Xie.
\newblock High performance offline handwritten chinese character recognition
  using googlenet and directional feature maps.
\newblock In {\em ICDAR}, pages 846--850. IEEE, 2015.

\end{thebibliography}

\end{document}